\documentclass[10pt,twocolumn,letterpaper]{article}

\usepackage[pagenumbers]{cvpr} % To force page numbers, e.g. for an arXiv version

\usepackage[dvipsnames]{xcolor}

\definecolor{cvprblue}{rgb}{0.21,0.49,0.74}
\usepackage[pagebackref,breaklinks,colorlinks,citecolor=cvprblue]{hyperref}
\usepackage{booktabs}
\usepackage{multirow}
\usepackage{tabularx}
\usepackage[capitalize]{cleveref}

\def\paperID{13} % *** Enter the Paper ID here
\def\confName{CVPR-W}
\def\confYear{2024}

\title{On the low-shot transferability of [V]-Mamba}

\author{Diganta Misra \thanks{equal contribution}\\
Mila - Quebec AI Institute, Landskape AI\\
{\tt\small diganta.misra@mila.quebec}
\and
Jay Gala \footnotemark[1]\\
AI4Bharat
\and
Antonio Orvieto\\
ELLIS Institute Tübingen, MPI for Intelligent Systems, Tübingen AI Center
}
\begin{document}
\maketitle
\begin{abstract}
The strength of modern large-scale neural networks lies in their ability to efficiently adapt to new tasks with few examples. Although extensive research has investigated the transferability of Vision Transformers (ViTs) to various downstream tasks under diverse constraints, this study shifts focus to explore the transfer learning potential of [V]-Mamba. We compare its performance with ViTs across different few-shot data budgets and efficient transfer methods. Our analysis yields three key insights into [V]-Mamba's few-shot transfer performance: (a) [V]-Mamba demonstrates superior or equivalent few-shot learning capabilities compared to ViTs when utilizing linear probing (LP) for transfer, (b) Conversely, [V]-Mamba exhibits weaker or similar few-shot learning performance compared to ViTs when employing visual prompting (VP) as the transfer method, and (c) We observe a weak positive correlation between the performance gap in transfer via LP and VP and the scale of the [V]-Mamba model. This preliminary analysis lays the foundation for more comprehensive studies aimed at furthering our understanding of the capabilities of [V]-Mamba variants and their distinctions from ViTs.
\end{abstract}
\section{Introduction}
\label{sec:intro}

\begin{figure}[t]
  \centering
  \includegraphics[width=0.99\linewidth]{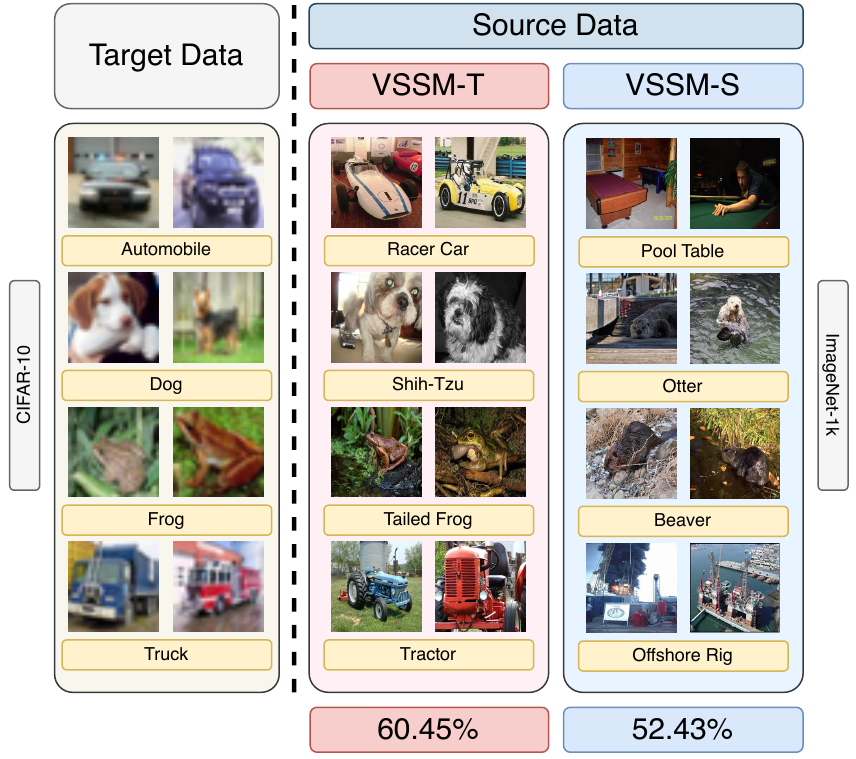}
  \caption{Comparison of label mappings from the source dataset (ImageNet-1k \citep{deng2009imagenet}) between VSSM-Tiny (VSSM-T) \cite{liu2024vmamba} and VSSM-Small (VSSM-S) \citep{zhu2024vision} models when transferred via ILM-VP \cite{chen2023understanding} for target classes in the CIFAR-10 \cite{krizhevsky2009learning} dataset. VSSM-Tiny demonstrates a more semantically accurate label mapping from the source dataset, while VSSM-Small associates target classes with semantically unrelated classes from the source dataset. Furthermore, the test accuracy at the bottom confirms the superiority of VSSM-Tiny over VSSM-Small.}
  \label{fig:analysis}
\end{figure}

Transfer learning~\cite{zhuang2020comprehensive,raffel2020exploring,neyshabur2020being} continues to be a key area of investigation within contemporary deep learning, despite the emergence of expansive foundational models~\cite{brown2020language,radford2021learning,team2023gemini,bommasani2021opportunities} endowed with robust zero-shot capabilities. Although pretraining methodologies traditionally rely on large datasets~\cite{schuhmann2022laion,sharma-etal-2018-conceptual}, this presumption may not align with the realities of downstream tasks, which often face constraints such as limited computational resources and data availability. Consequently, there is a pressing need to develop methodologies tailored to the efficient transfer of knowledge in few-shot learning scenarios.

A plethora of efficient transfer learning techniques have been proposed, spanning a spectrum from parametrically efficient approaches such as Low Rank Adaptation (LoRA)~\cite{hu2021lora} and Adapters~\cite{houlsby2019parameter,sung2022vl} to more conventional methods such as Linear Probing (LP)~\cite{alain2016understanding} and Full Fine-tuning (FF). Drawing inspiration from advances in language models, the concept of prompting, in the form of Visual Prompting (VP)~\cite{chen2022model,elsayed2018adversarial,bahng2022exploring,tsai2020transfer,zhang2022fairness,salman2021unadversarial,vinod2023reprogramming,yang2021voice2series}, has garnered considerable attention as a cheap and effective means of tailoring robust pre-trained vision models to specific downstream tasks.

However, regarding efficiency and domain-specific applicability, both Linear Probing (LP) and Visual Prompting (VP) persist as formidable contenders for transfer modes. Extensive scrutiny~\cite{misra2023reprogramming,chen2023understanding} has been directed towards these methodologies within the realm of Convolutional Neural Networks (CNN) and Vision Transformers (ViTs)~\cite{dosovitskiy2020image}. Nevertheless, with the advent of Mamba-based models tailored for visual perception~\cite{liu2024vmamba,zhu2024vision}, the efficacy of such State Space Model (SSM) variants compared to Transformers regarding downstream adaptation remains an open question awaiting empirical exploration.

Thus, the primary objective of this study is to conduct an empirical assessment of the low-shot transfer capabilities exhibited by variants of Visual Mamba ([V] -Mamba) compared to Vision Transformers (ViT), with a specific focus on elucidating the performance differential observed between transfer mechanisms utilizing Linear Probing (LP) and Visual Prompting (VP) methods.

Outlined below are the key findings derived from our analysis.

\begin{enumerate}
    \item \emph{[V]-Mamba are stronger few-shot learners than ViTs when transferred via LP.}
    \item \emph{[V]-Mamba are weaker few-shot learners than ViTs when transferred via VP.}
    \item \emph{A weak positive correlation emerges between the performance gap noted in transferring via Linear Probing (LP) and Visual Prompting (VP) methods and the increasing scale of the [V]-Mamba model.}
\end{enumerate}

\section{Related Work}
\label{sec:rel_work}

\subsubsection*{State-space models and Mamba}
State-space models sparked from the seminal S4 work by~\citet{gu2021efficiently}, who leveraged insights from recursive signal estimation theory~\citep{gu2020hippo} to design a deep transformer-like model~\citep{vaswani2017attention} where attention is replaced by a carefully parameterized linear recurrent neural network. The design of S4 got drastically simplified in~\citep{gupta2022diagonal, gu2022parameterization, gupta2022simplifying}, who first achieved state of the art on the long-range arena~(LRA)~\citep{tay2020long} with a highly efficient linear diagonal recurrent mechanism, where forward and backward passes leverage a convolutional view on SSMs~\citep{gu2021combining,li2022makes}, or parallel scans on GPU~\citep{martin2017parallelizing, smith2023simplified, orvieto2023resurrecting}. 

The efficient recurrent~(therefore linear in sequence length) nature of SSMs makes them particularly appealing when compared to attention-based transformers, where both inference time and memory suffer quadratically from sequence length. In addition to the LRA benchmark, SSMs found first successful applications in vision~\citep{nguyen2022s4nd}, audio~\citep{goel2022sashimi} as well as online learning~\citep{zucchet2023online}/reinforcement learning~\citep{lu2023structured}. Initial attempts in language modeling~\citep{fu2022hungry, wang2022pretraining}, as well as theoretical investigations~\citep{orvieto2024universality, wang2023state} gave further insights on necessary architectural improvements unlocking the NLP domain. Leveraging deep connections between the gating and attention, a few works~\citep{peng2023rwkv, sun2023retentive, katsch2023gateloop} started incorporating input selectivity mechanisms~(a crucial feature for in-context learning~\citep{olsson2022context}) into SSMs. These efforts culminated in the Mamba architecture~\citep{gu2023mamba}, which smartly combines input selectivity with MLP-gating of the selective SSM layer, improving on previous solutions proposed in H3~\citep{fu2022hungry} and Hyena~\citep{poli2023hyena} on text. The design of Mamba is also strongly supported by theoretical evidence that shows its superior expressive power compared to S4~\citep{cirone2024theoretical}, and its close relation to softmax attention~\citep{ali2024hidden}. 

Beyond text, Mamba was recently applied to the vision domain, yielding two architectures  --VMamba~\citep{liu2024vmamba} and Vim~\citep{zhu2024vision} -- with outstanding results compared to ViTs~\citep{Touvron2022DeiTIR} both in terms of performance~(e.g. ImageNet-1K top-1 accuracy) and efficiency~(inference time, memory, parameter efficiency). At the core of this success is the sequential mechanism of Mamba, adapted to image data through a simple bidirectional processing of patches relying on positional embeddings~\citep{zhu2024vision} or a more sophisticated Cross-scan~\cite {liu2024vmamba} technique unfolding image patches along rows and columns into sequences, and then proceeding along four directions. 

\subsubsection*{Visual Prompting and Transferability}
Visual prompting introduces a parameter-efficient fine-tuning method by integrating input transformation and output mapping layers into a pre-trained model for transfer learning \cite{chen2022model,elsayed2018adversarial}. VP achieves input transformation through trainable additive padding operations for each image and output mapping by transitioning from source to target label classes. Various output mapping strategies have been proposed, including frequency-guided label mapping \cite{tsai2020transfer} and iterative label mapping (ILM-VP) \cite{chen2023understanding}. Previous research has also explored the possible adverse effects associated with the use of Visual Prompting (VP) as the transfer mode for compressed models \cite{misra2023reprogramming}. 

The literature on transfer learning spans a wide spectrum, encompassing studies that explore various domains, such as art \cite{sabatelli2018deep}, as well as more theoretically oriented investigations focusing on inductive biases in transfer learning \cite{xuhong2018explicit}. In addition, researchers have explored alternative approaches, ranging from leveraging reinforcement learning for adaptive transfer \cite{zhu2020learning} to conventional adaptive fine-tuning strategies \cite{guo2019spottune,guo2020adafilter}. 

In a study by Liu et al. \cite{liu2019towards}, the impact of pre-trained representations on transfer environments was examined, demonstrating how transferring pre-trained representations can aid in identifying superior and more stable minima in the objective landscape. In particular, Donahue et al. \cite{donahue1deep} initially proposed the use of a new probe to classify features extracted by a pre-trained model at various depths of the AlexNet architecture \cite{krizhevsky2009learning}.

Moreover, Oquab et al. \cite{oquab2014learning} showcased the significant benefits of transferring pre-trained features obtained from image classification tasks to downstream tasks with dense annotations, such as object detection.

\section{Experimental Setup}

In this section, we provide in-depth details on the extensive experiments and their configuration, including datasets, models, and training hyperparameters.

\subsection{Datasets}\label{subsec:datasets}

We consider seven datasets that encompass a mixture of downstream tasks across both near and far domains. These datasets include CIFAR-10 \cite{krizhevsky2009learning}, SVHN \cite{netzer2011reading}, GTSRB \cite{Houben-IJCNN-2013}, DTD \cite{cimpoi14describing}, Flowers-102 \cite{Nilsback08}, OxfordPets \cite{parkhi12a}, and EuroSAT \cite{helber2018introducing}. CIFAR-10 consists of 50,000 training images and 10,000 test images, each 32 x 32, divided into 10 classes representing various objects and animals. SVHN comprises 73,257 training images and 26,032 test images, each of size 32 x 32, categorized into 10 classes depicting digits from Google Street View images. GTSRB consists of 39,209 training images and 12,630 test images, categorized into 43 classes that showcase various traffic signs. DTD contains 1,880 training images and 1,880 test images, each of size ranging between 300 x 300 to 640 x 640 pixels, covering 47 texture classes. Flowers-102 includes 2,040 training images and 6,149 test images, spanning 102 distinct categories of flower species. OxfordPets consists of 3,680 training images and 3,669 test images, representing 37 different pet breeds. EuroSAT comprises 24,300 training images and 2,700 test images, each sized at 64 x 64 pixels, representing 10 land use and land cover categories commonly found in European satellite imagery.

\subsection{Models}\label{subsec:models}

We base our experiments on models pre-trained on the ImageNet-1k \cite{deng2009imagenet} classification task, categorizing them into two families of model architectures, namely ViTs and SSMs. We use three standard vision classifiers following ViT: (a) DeiT-Small \cite{pmlr-v139-touvron21a}, (b) DeiT3-Small \cite{Touvron2022DeiTIR} and (c) MoCov3-Small \cite{chen2021empirical}. On the other hand, we use two SSM-based classifiers: (a) VSSM-Tiny \cite{liu2024vmamba} and (b) Vim-Small \cite{zhu2024vision}. We specifically choose small and tiny models to ensure comparable model sizes across two model families for fair comparison across different experimental settings. In addition, we also experimented with the other scale variants of Vim and VSSM, specifically Vim-Tiny \cite{zhu2024vision} and VSSM-Small \cite{liu2024vmamba}.

\subsection{Training Details}

We investigate the low-shot transferability of different models across different datasets described in \Cref{subsec:models,subsec:datasets} with two methods: linear probing (LP) and visual prompting (VP). For the VP method, we employ state-of-the-art Iterative Label Mapping (ILM-VP) \cite{chen2023understanding}. We train all models for 100 epochs using the Adam optimizer \cite{kingma2014adam}. We employ a multistep learning rate decay scheduler that reduces the initial learning rate of 0.01 by a factor of $\frac{1}{10}$ at the 50th and 72nd epochs, respectively. \Cref{tab:batch_size} list the specific batch sizes used for training with each N-shots configuration. ``N-shots” refers to the training data budget, i.e. the number of samples per class of the downstream dataset used during training. For experiments employing ILM-VP, the images are resized to 32 x 32 for the \textup{CIFAR-10}, \textup{GTSRB}, and \textup{SVHN} datasets, while a 128 x 128 size version is utilized for the other datasets. Conversely, LP experiments are conducted with images resized to 224 x 224 across all datasets. All experiments were implemented using the PyTorch framework \cite{pytorch} and the Timm library \cite{rw2019timm} and performed on a single A100 40Gb GPU. We run all configurations with three seeds to ensure consistency and measure statistical significance.

\begin{table}[t]
\centering
\begin{tabular}{l|cccc}
     \textbf{N-shots} & 1 & 10 & 50 & 100 \\
     \midrule
     \textbf{Batch size} & 8 & 32 & 64 & 128 \\
\end{tabular}
\caption{The different batch sizes used for each N-shots configuration for all experiments configurations.}
\label{tab:batch_size}
\end{table}

\section{Results}

\begin{figure*}[t]
  \centering
  \includegraphics[width=0.97\linewidth]{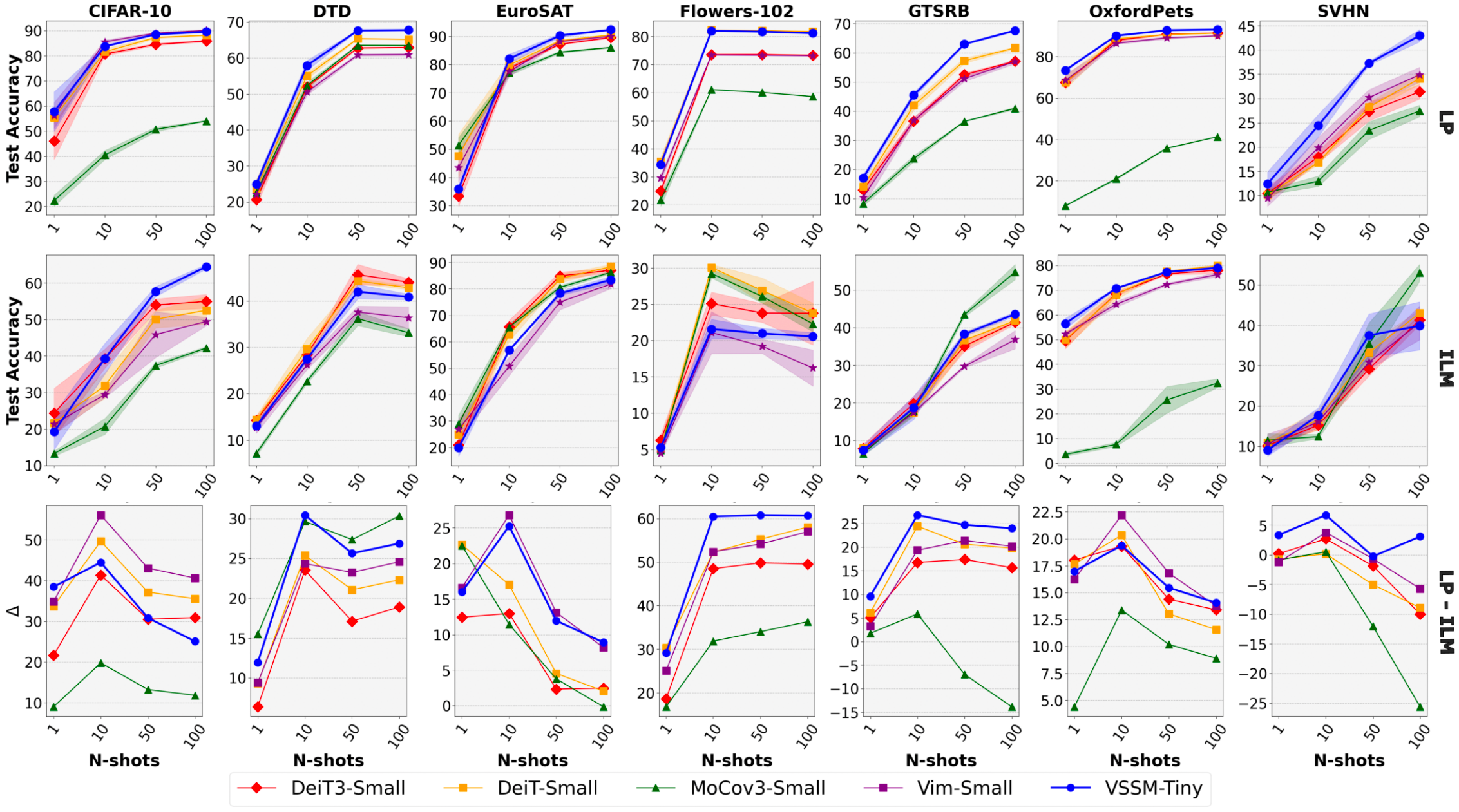}
  \caption{Transfer performance measured by test accuracy of different models of similar scales across various downstream datasets at different $N$-shot settings trained using LP (top) and ILM-VP \citep{chen2023understanding} (middle) method. $\Delta$ (bottom) denotes the difference in test accuracy between LP and ILM-VP models across varying datasets and data budgets.}
  \label{fig:main_res}
\end{figure*}

\begin{figure*}[t]
  \centering
  \includegraphics[width=0.97\linewidth]{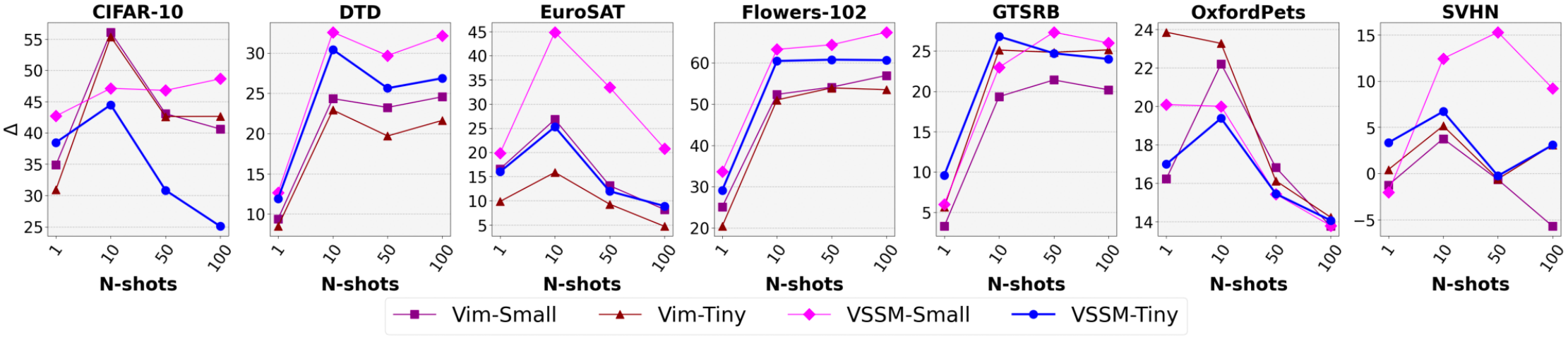}
  \caption{Transfer performance gap ($\Delta$) measured by the difference in test accuracy between LP and ILM-VP \citep{chen2023understanding} methods for various SSM models on a variety of downstream datasets at different $N$-shot settings.}
  \label{fig:delta}
\end{figure*}

In this section, we analyze the experimental findings with the aim of understanding the transferability of LP and ILM-VP across two model families, namely ViTs and SSMs, specifically under low data volume constraints. 

\subsubsection*{[V]-Mamba are better or same as ViTs for LP}

In \Cref{fig:main_res} for LP, we see that, in general, SSM models consistently demonstrate superior or comparable performance to various ViT variants of comparable scales across diverse datasets, encompassing data budgets ranging from 1 to 100 shots. MoCov3-Small, which is a ViT variant, is outperformed by a significant margin by the SSM counterparts, Vim-Small and VSSM-Tiny across most datasets, with the exceptions of DTD and EuroSAT datasets where comparable performance is observed. Furthermore, performance trends tend to stabilize beyond 10-shots across both ViT and SSM variants as observable in datasets such as CIFAR-10, EuroSAT, Flowers-102 and OxfordPets. Although DeiT3-Small outperforms the DeiT-Small on ImageNet-1k pretraining, we do observe that the DeiT-Small slightly outperforms DeiT3-Small for a few datasets such as GTSRB and DTD with increasing shots.

\vspace{-3mm}

\subsubsection*{[V]-Mamba are worse or same as ViTs for VP}

In \Cref{fig:main_res} for ILM-VP, we find a trend contrary to that of LP, where SSM models perform poorly or are comparable in performance compared to their ViT counterparts. Among the SSM models, Vim-Small underperforms in comparison to VSSM-Tiny across multiple datasets including CIFAR-10, DTD, Flowers-102, and GTSRB, particularly at higher shot settings. Moreover, we observe varying performance trends for MoCov3, a ViT variant, across different datasets and data budgets under consideration. In the case of the OxfordPets dataset, MoCov3 performs the worst, with the performance gap exceeding 50\% consistently for different data budgets compared to other models. Additionally, we see a decreasing trend in performance for the Flowers-102 dataset with higher shots across both ViT and SSM variants. Similarly, a slight decline in performance is evident for the 100-shot data budget setting on the DTD dataset across all models under consideration.

\vspace{-4mm}

\subsubsection*{On the weak positive correlation between performance disparity and progressive scaling of [V]-Mamba}

In \Cref{fig:main_res}, the bottom subplot illustrates the performance trends for the difference in test accuracy ($\Delta$) between the LP and ILM-VP methods for different models across various datasets under varying levels of data budgets. We can infer that $\Delta$ is the highest for VSSM-Tiny followed by Vim-Small for most of the datasets highlighting that the effectiveness of VP falls short compared to LP in terms of transferability on downstream datasets. Overall, we see that $\Delta$ tends to decrease as the number of shots increases for 4 datasets out of 7 datasets except in the case of the 10-shot setting where it appears to be elevated. This suggests that the VP method seems to close the performance gap at higher shots.

In \Cref{fig:delta}, we report the performance trends for the difference in test accuracy ($\Delta$) between the LP and ILM-VP methods for varying scales of SSM models across different datasets and data budgets. We can see that, in general, there appears to be a consistent pattern where $\Delta$ increases as the model size increases, as is evident in both SSM variants, Vim and VSSM. Furthermore, we observe that the trends for VSSM-Tiny and VSSM-Small are distinctively separated for most of the datasets whereas Vim-Tiny and Vim-Small closely follow each other for most of the datasets. This finding warrants further exploration of ways to mitigate the transferability gap between the LP and VP methods with the progressive scaling of the SSMs.

\section{Conclusion}

This study aims to investigate the few-shot transfer efficacy of [V]-Mamba compared to Vision Transformers (ViTs) across diverse downstream classification datasets. Two efficient transfer methodologies, namely linear probing and visual prompting, are employed for this analysis. Our results indicate that [V]-Mamba demonstrates either superior or comparable few-shot learning capabilities to ViTs when transferred via linear probing. Conversely, when employing visual prompting for few-shot transfer, [V]-Mamba's performance tends to be weaker or equivalent to ViTs. Furthermore, we observe a weak positive correlation between the performance gap in transfer via linear probing and visual prompting and the increasing scale of the [V]-Mamba model. We anticipate that our study will lay the groundwork for further exploration aimed at comprehensively elucidating the capabilities of [V]-Mamba variants relative to those of ViTs.

{
    \small
    \bibliographystyle{ieeenat_fullname}
    \bibliography{main}
}


\end{document}